\documentclass[journal,twoside,web]{ieeecolor}
\usepackage{generic}
\usepackage{cite}
\usepackage{amsmath,amssymb,amsfonts}
\usepackage{algorithmic}
\usepackage{booktabs}
\usepackage{graphicx}
\usepackage{algorithm,algorithmic}
\usepackage{multirow}
\usepackage{hyperref}
\hypersetup{hidelinks=true}
\usepackage{textcomp}
\def\BibTeX{{\rm B\kern-.05em{\sc i\kern-.025em b}\kern-.08em
    T\kern-.1667em\lower.7ex\hbox{E}\kern-.125emX}}
\begin{document}
\title{Agent-Driven Generative Priors for Pathology Image Synthesis and Segmentation}
\author{Sicheng Yang, Rui Tang, Haipeng Zhou, Hongqiu Wang, Zhaohu Xing, Jianda Ma, Lie Dai, \\ Guang Yang, and Lei Zhu
\thanks{Sicheng Yang, Haipeng Zhou, Hongqiu Wang, and Zhaohu Xing are with the Hong Kong University of Science and Technology (Guangzhou), China.  
}
\thanks{Rui Tang is with The University of Hong Kong,
Pokfulam, Hong Kong SAR, China.}
\thanks{Jianda Ma and Lie Dai are with Department of Rheumatology and Immunology, Sun Yat-Sen Memorial Hospital, Sun Yat-Sen University, China.}
\thanks{Guang Yang is with Imperial College London, UK.}
\thanks{Lei Zhu is with The Hong Kong University of Science and Technology (Guangzhou), China, and The Hong Kong University of Science and Technology, Hong Kong SAR, China.}
\thanks{Lei Zhu (E-mail: leizhu@hkust-gz.edu.cn) is the corresponding author of this work.}
}

\maketitle

\begin{abstract}
Deep learning methods for pathology are constrained by the limited availability and high cost of large-scale pixel-level annotations. Synthetic data can alleviate this limitation, but existing pipelines often lack reliable quality control and remain weakly connected to downstream analysis. To address these issues, we propose a three-stage framework that integrates automated data screening, generative representation learning, and efficient adaptation to downstream tasks. First, we develop a pathology agent system to assess image quality and filter samples, resulting in TCGA-200K, a dataset containing 217,030 pathology patches selected from the public TCGA repository. Second, we train a generative model on the screened data. In addition to image synthesis, its backbone learns structural and semantic representations that can be transferred to downstream tasks. Third, we adapt the pretrained generative backbone to semantic segmentation using a lightweight task-specific head. We further introduce PathSeg-380K, a semantic segmentation dataset with expert annotations for approximately 380,000 cells across 12 cell types. Experiments on multiple pathology datasets show that the proposed framework achieves state-of-the-art segmentation performance. 
Our code is publicly available at \url{https://github.com/script-Yang/Agent-GenPath}.
\end{abstract}


\section{Introduction}
\label{sec:introduction}
\begin{figure}[t]
    \centering               
\includegraphics[width=0.46 \textwidth]{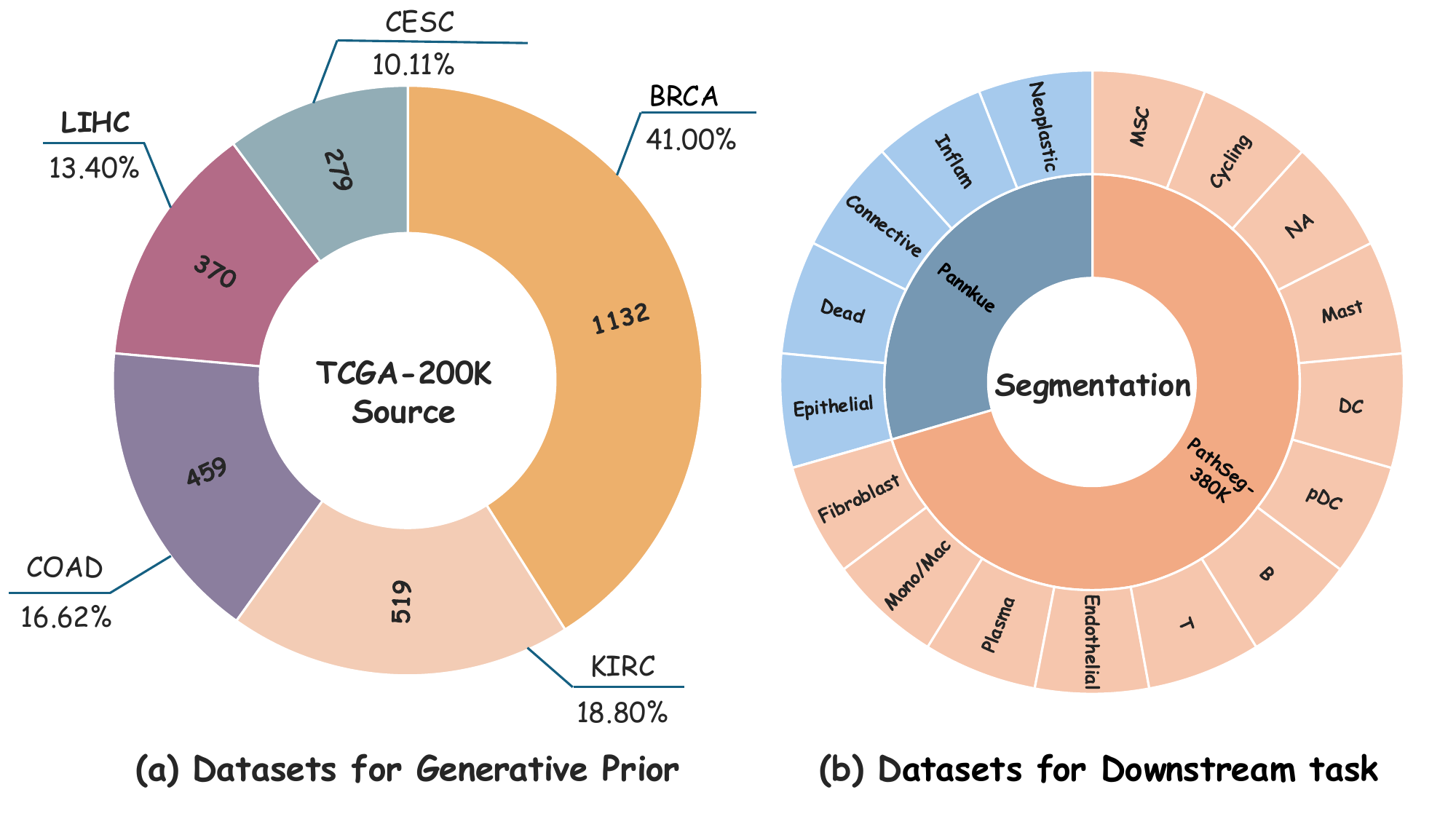}
    \caption{Overview of the datasets and data distributions. (a) Source composition of TCGA-200K, utilized for pre-training the generative prior. (b) Class distributions for downstream segmentation tasks, comparing the publicly available PanNuke dataset with our large-scale, annotated PathSeg-380K dataset, containing $\sim$ 380,000 annotated cells.}
    \label{fig:dataset}
    \vspace{-5mm}
\end{figure}
\IEEEPARstart{H}istopathological examination is a cornerstone of clinical diagnosis and disease assessment~\cite{yang2026lcm}. Advances in deep learning have further expanded the capabilities of computational pathology, supporting applications ranging from nuclei segmentation~\cite{verma2021monusac2020} and tumor detection~\cite{ehteshami2017diagnostic} to tissue classification~\cite{coudray2018classification}. However, these models often rely on large-scale expert annotations~\cite{puttapirat2018openhi,wang2024cytogan,zhou2025coc}. Obtaining such annotations is costly and time-consuming, as accurate labeling requires pathologists to manually identify and delineate complex cellular structures~\cite{schaumberg2018large}. As a result, sufficiently annotated datasets remain scarce in many clinical settings~\cite{sylolypavan2023impact}, increasing the risk of overfitting and limiting model generalization.
\begin{figure*}[t]
    \centering               
\includegraphics[width=1.0\textwidth]{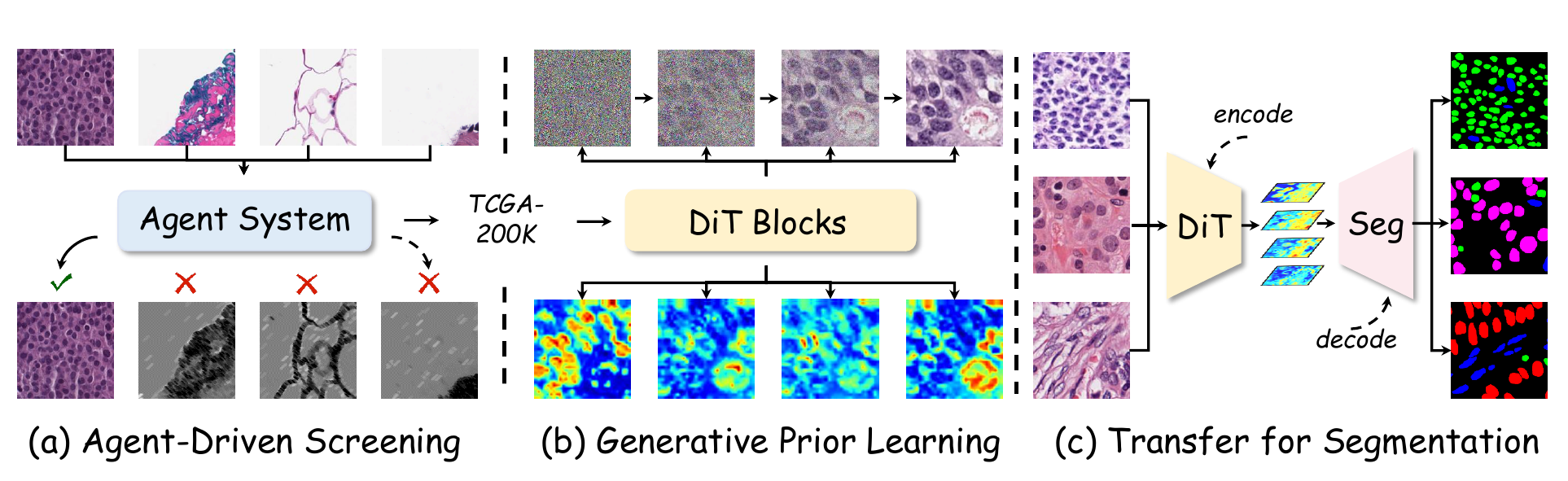}
    \caption{Overview of the proposed unified three-stage framework. In Stage I, an agent-driven screening pipeline evaluates pathology patches from multiple quality dimensions and selects high-quality samples to construct the TCGA-200K dataset. In Stage II, the curated data are used to learn pathology-specific generative priors through DiT blocks. In Stage III, the learned generative priors are transferred to downstream segmentation on PanNuke and PathSeg-380K, where the pretrained backbone provides multi-level pathology representations for accurate pathology segmentation.}
    \label{fig:main}
\end{figure*}

Synthetic data generation has emerged as a promising strategy to augment scarce labeled data~\cite{yi2019generative}. Generative models, including generative adversarial networks (GANs)~\cite{goodfellow2020generative} and diffusion models~\cite{ho2020denoising}, have demonstrated strong capability in synthesizing realistic pathology images. Typically, these generated images are combined with the existing datasets to augment the training pool, thereby improving the performance of downstream tasks such as segmentation~\cite{yu2023diffusion}. However, it still faces two major limitations. (1) Existing pipelines lack effective data quality control. In the pathology area, images frequently suffer from staining artifacts, blurred structures, and poor cellular visibility~\cite{harb2024diffusion,yang2025k}. For example, within a single whole slide image (WSI), many regions are often uninformative for diagnosis. Training generative models on such low-quality data without quality assessment leads to unreliable synthetic images with inaccurate pathological patterns~\cite{brodsky2025generative}. (2) The generation process and downstream tasks are disconnected from each other. Existing methods treat generative models merely as black-box data engines to sample new images. Therefore, they fail to leverage the rich semantic information learned during the generation process~\cite{zhang2021datasetgan}. Our experiments reveal that the generative model backbone itself can serve as a powerful feature extractor for downstream tasks, providing rich prior information.

To address these limitations, we propose a novel three-stage framework that unifies automated data quality control, the learning of high-quality generative priors, and transfer learning for downstream tasks. First, we introduce an agent-based screening mechanism for dataset refinement. We design an intelligent pathology agent system to automatically evaluate and score the quality of pathology images across multiple clinical dimensions. Only high-scoring images are retained to construct a curated training set for learning the generative prior. Second, we train a generative model on this refined dataset. Crucially, this model is not deployed merely to synthesize new data; instead, its backbone naturally extracts rich structural and semantic information during the generative training process, allowing it to serve as a powerful foundational encoder. Third, we redefine downstream analysis as a transfer learning paradigm. By simply appending a lightweight task-specific head (e.g., for semantic segmentation) to the pre-trained generative backbone, our framework leverages these learned priors with minimal training overhead. Our contributions are summarized as follows:
\begin{itemize}
\item \textbf{A Unified Three-Stage Framework.} We identify that current pathology analysis pipelines suffer from inadequate data quality control and a disconnect between generation and downstream tasks. To address this, we propose a novel framework that unifies agent-guided quality screening, robust generative prior learning, and efficient downstream transfer learning.
\item \textbf{Two Novel High-Quality Datasets.} We construct and introduce two large-scale datasets to the community: TCGA-200K, comprising 217,030 premium pathology patches curated from the open-source TCGA repository using our agent-based screening mechanism. We also use PathSeg-380K for semantic segmentation. It contains annotations provided by two clinical pathologists for approximately 380,000 cells across 12 cell types.

\item \textbf{State-of-the-Art Performance.} Extensive experiments show that our framework generates high-quality pathology images. When the pretrained generative backbone is transferred to semantic segmentation, the resulting model also outperforms existing segmentation methods across both public and private datasets, indicating that the learned generative prior benefits both image synthesis and downstream pathology analysis.
\end{itemize}
\subsection{Synthetic Data for Pathology}
Acquiring massive and fully annotated pathology datasets is frequently constrained by privacy concerns and high costs~\cite{xiao2025artificial}. To address this data scarcity, researchers have actively explored synthetic data generation~\cite{yellapragada2025pixcell}. While models such as Generative Adversarial Networks (GANs) and diffusion models have achieved notable progress in producing visually realistic images~\cite{ali2025generative,yang2023diffusion}, a critical limitation persists. Current approaches largely neglect the rigorous filtering of data sources~\cite{zhang2024generate,weng2024grandqc}. Consequently, incorporating training images of poor quality inevitably introduces erroneous information into the generative models. 
As a result, the synthesized images may fail to preserve key pathological structures, reducing their usefulness for downstream medical tasks.

\subsection{Agents in Pathology}
The reasoning capabilities of AI agents have opened new frontiers in medical image analysis~\cite{sun2025cpathagent}. These agents have shown remarkable potential in automating complex diagnostic workflows, including pathological image interpretation, report generation, and interactive clinical decision-making~\cite{li2024agent}. By simulating the reasoning processes of human experts, agents can analyze intricate tissue morphologies and cellular patterns. Despite their growing success in diagnostic tasks, the ability of agents to guide pathological image generation remains largely unexplored~\cite{ghezloo2025pathfinder}. Specifically, they can act as virtual experts, providing feedback on image quality and structural validity, which offers a promising avenue for enhancing the reliability of synthetic medical data. Motivated by this gap, we introduce a novel framework that leverages agentic reasoning to screen and guide pathological image synthesis.

\subsection{Pathology Image Segmentation}
Semantic segmentation is widely used in computational pathology to delineate cellular structures and tissue regions~\cite{zhou2019unet++,zhang2023sam,danisetty2025pathsegdiff,chen2026beyond}. Some methods train segmentation networks from scratch on task-specific datasets, but limited pixel-level annotations often lead to overfitting~\cite{komura2023restaining}. Others adopt backbones pretrained on natural images, such as ResNet~\cite{he2016deep} and DINO~\cite{caron2021emerging,yang2025segdino,yang2026segdino,yang2026vq}, whose representations do not fully capture pathology-specific morphology~\cite{kataria2023pretrain}. Pathology foundation models provide stronger domain-specific features~\cite{lu2024visual,chen2024towards}, yet they are designed as encoders and cannot support image generation. We instead pretrain a generative model on agent-curated pathology images and transfer its multi-level features to semantic segmentation, allowing the same generative prior to serve both image synthesis and downstream analysis.
\section{Methodology}
\begin{figure}[t]
    \centering               
\includegraphics[width=0.46 \textwidth]{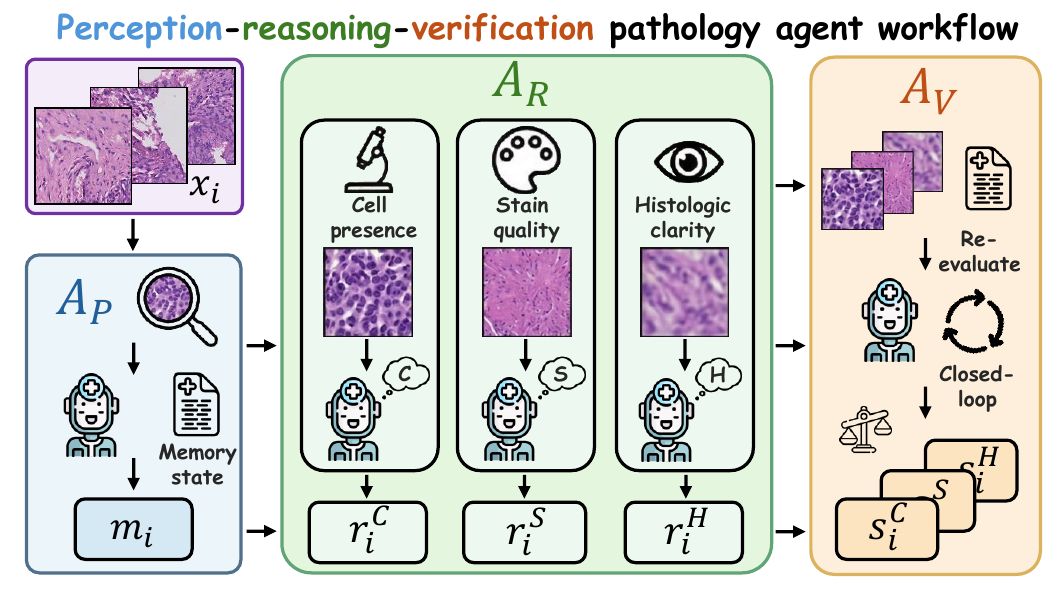}
    \caption{Stage I: Agent-driven high-quality pathology image selection. Given an input pathology patch $x_i$, $\mathcal{A}_P$ initializes the memory state $m_i$, $\mathcal{A}_R$ evaluates cell presence, stain quality, and histologic clarity, and $\mathcal{A}_V$ performs closed-loop verification to produce the final quality scores.}
    \label{fig:agent}
\end{figure}
\subsection{Overview}
As illustrated in Figure~\ref{fig:main}, we propose a unified three-stage framework that tightly integrates automated data quality control, generative prior learning, and downstream segmentation. First, we introduce an agent-driven screening mechanism to evaluate and filter raw pathology patches, yielding a highly curated, large-scale dataset (TCGA-200K). Second, we leverage this refined dataset to train a Diffusion Transformer (DiT) via Flow Matching, enabling the backbone to learn robust structural and semantic generative priors rather than serving merely as a data synthesizer. Finally, we reformulate downstream semantic segmentation as a transfer learning paradigm, where a lightweight multi-level decoder is appended to the pre-trained DiT backbone to map the extracted multi-scale features to precise pixel-level predictions. The effectiveness and generalization ability of this unified pipeline are comprehensively validated on both the public PanNuke dataset and our newly proposed PathSeg-380K dataset.

\subsection{Stage I: Agent-Driven Pathology Data Curation}
Current generative modeling in pathology is constrained by two major challenges: the limited scale of existing datasets, which often contain only a few thousand images, and the lack of rigorous quality control. As a result, low-quality samples, such as patches with sparse cellular content or out-of-focus blur, can substantially degrade model performance. To address this, we propose an agent-driven and scalable quality curation pipeline with a \textbf{semi-supervised} design. Specifically, we first construct a large-scale pathology dataset, denoted as $\mathcal{D}$, from multiple public TCGA cohorts~\cite{tomczak2015review}, and partition it into a small labeled pilot set $\mathcal{D}_P$ and a large unlabeled pool $\mathcal{D}_U$. A Multi-dimensional Assessment Pathology Expert Agent is then employed to annotate $\mathcal{D}_P$ across three key quality dimensions, providing supervision for training a lightweight regressor. The trained regressor is subsequently used to score the much larger unlabeled pool $\mathcal{D}_U$, enabling scalable quality estimation in a semi-supervised manner. Finally, a multi-dimensional quantile filtering strategy is applied to extract a high-fidelity subset $\mathcal{D}^*$, resulting in \textbf{TCGA-200K}, a curated dataset of over 200,000 high-quality pathology patches for generative modeling.
\begin{figure}[t]
    \centering               
\includegraphics[width=0.46 \textwidth]{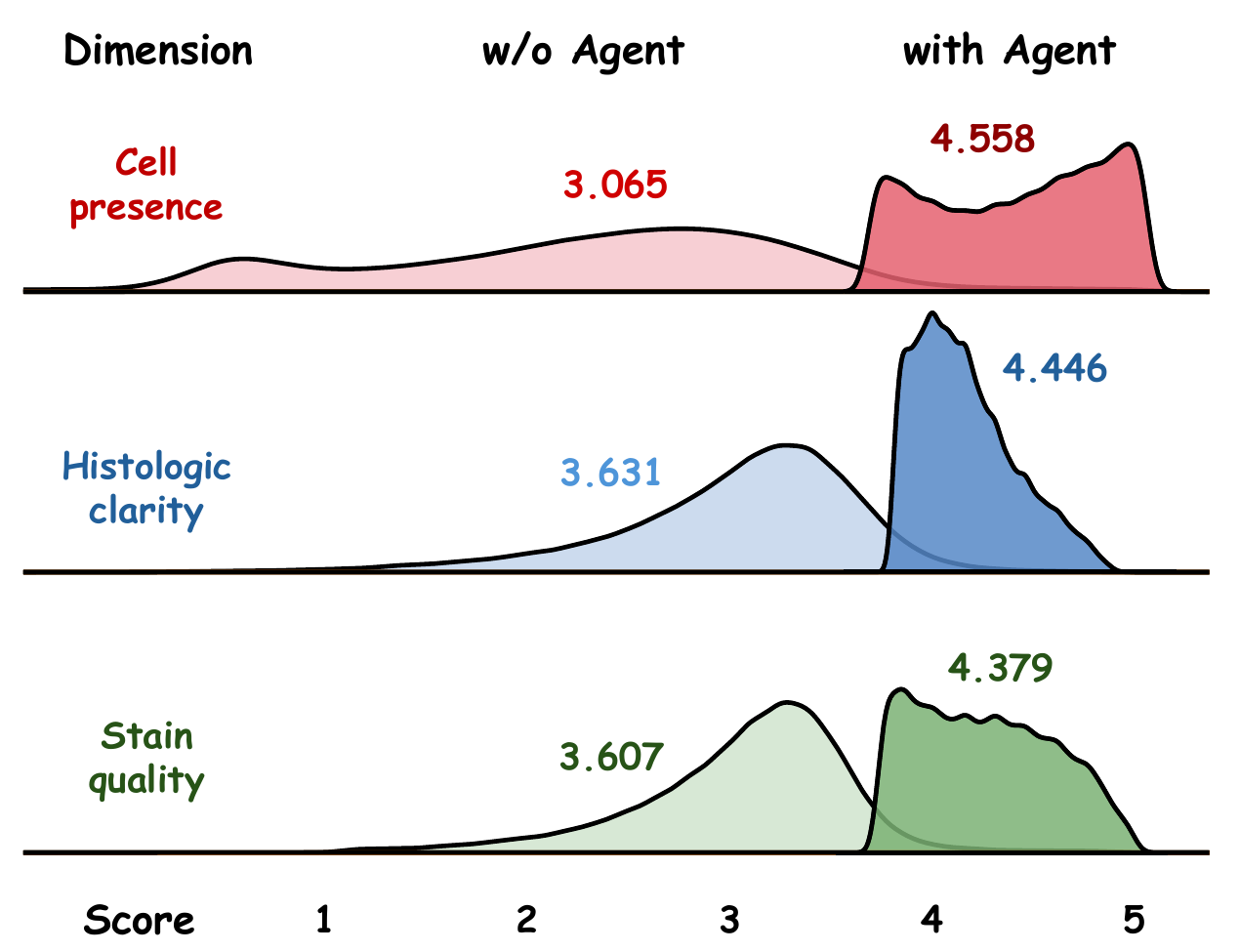}
\caption{Score distributions before and after the integration of our pathology expert agent across three quality dimensions. The distributions shift toward higher scores after agent-driven filtering, indicating an improvement in overall data quality.}
    \label{fig:path_score}
\end{figure}
\begin{figure*}[t]
    \centering               
\includegraphics[width=1.0 \textwidth]{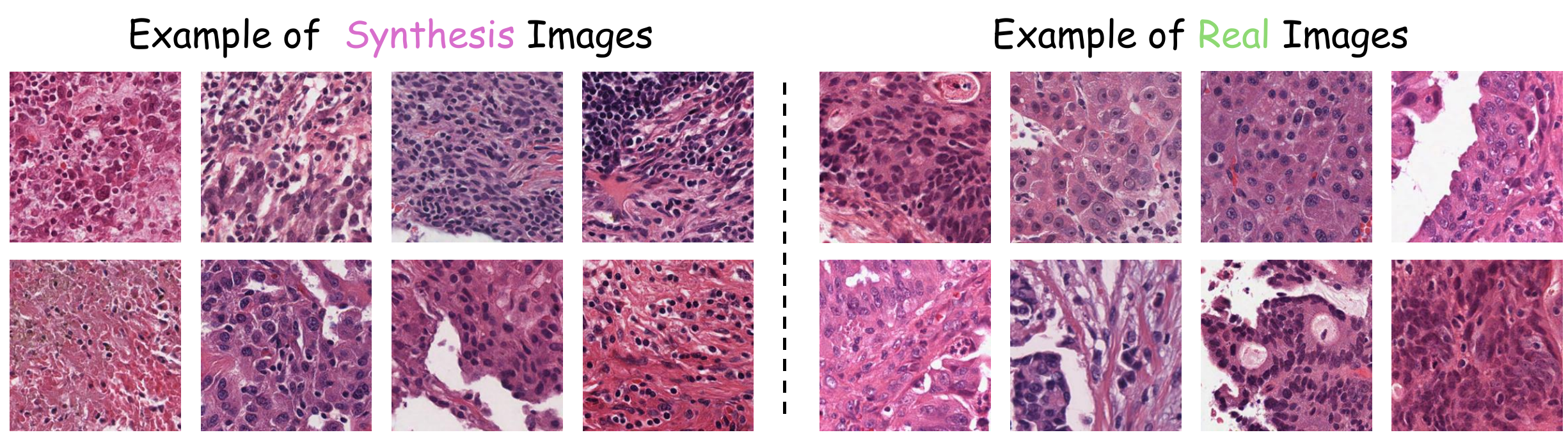}
\caption{Comparison of synthetic (left) and real (right) pathology images. The generated samples faithfully reproduce complex tissue morphology, cellular structures, and spatial organization, showing strong visual consistency with real clinical images.}
    \label{fig:fake_img}
\end{figure*}
\begin{figure}[t]
    \centering               
\includegraphics[width=0.46 \textwidth]{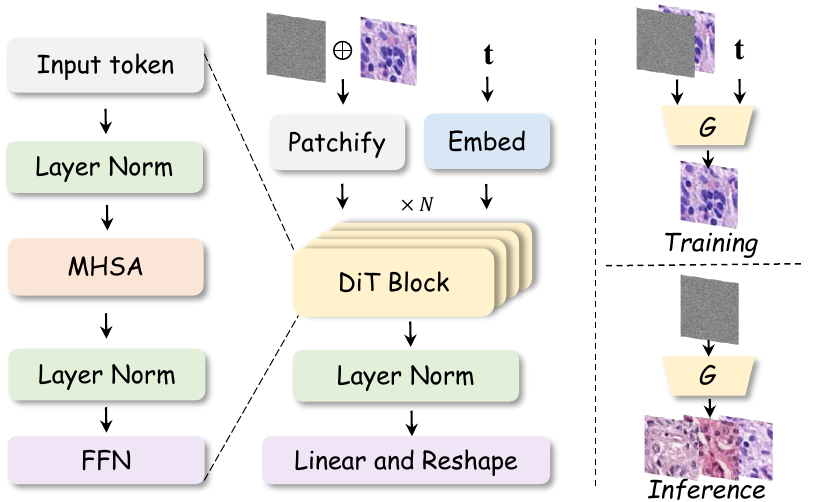}
        \caption{Stage II: Learning pathology generative priors via Flow Matching. Given an image patch and timestep $t$, the DiT generator $G$ is trained to predict the target flow field.}
    \label{fig:dit}
\end{figure}
\subsubsection{Large-Scale Pathology Dataset Construction}
As illustrated in Fig.~\ref{fig:dataset} (a), we source a diverse set of Whole Slide Images (WSIs) across five cancer types from TCGA: Breast Invasive Carcinoma (BRCA), Cervical Squamous Cell Carcinoma (CESC), Colon Adenocarcinoma (COAD), Kidney Renal Clear Cell Carcinoma (KIRC), and Liver Hepatocellular Carcinoma (LIHC). We employ the CLAM toolkit~\cite{lu2021data} to process these WSIs into a comprehensive dataset $\mathcal{D}$ containing approximately \textbf{14 million} $256 \times 256$ patches. While this preprocessing effectively removes the majority of the white background regions, the retained patches inevitably exhibit localized artifacts (e.g., tissue folding, out-of-focus blurring, and uneven staining) that can destabilize training dynamics. To facilitate our agent-driven quality curation, we randomly partition the full dataset $\mathcal{D}$ into two disjoint subsets: a small pilot set $\mathcal{D}_P$ and a massive unannotated target pool $\mathcal{D}_U$.

\subsubsection{Pathology Expert Agent for Quality Supervision}
To obtain reliable quality pseudo-labels without exhaustive manual annotation, we design a Multi-dimensional Assessment Pathology Expert Agent. The agent follows a structured \textit{Perception--Reasoning--Verification} workflow to evaluate the visual quality and pathological content of each patch in the pilot set $\mathcal{D}_P$.

As shown in Fig.~\ref{fig:agent}, for each image $x_i \in \mathcal{D}_P$, the agent's perception module $\mathcal{A}_P$ first conducts a global inspection to identify tissue types and prominent artifacts (\textit{e.g.}, folding or blurring). 
This preliminary step initializes a contextual memory state $m_i = \mathcal{A}_P(x_i)$ that guides subsequent decision-making. 
Conditioned on $m_i$, the reasoning module $\mathcal{A}_R$ then autonomously analyzes three pathologically critical dimensions: cell presence ($C$), stain quality ($S$), and histologic clarity ($H$). 
Instead of directly outputting ratings, the agent is mandated to explicitly generate a diagnostic rationale $r_i^d$ before taking a preliminary scoring action $\tilde{s}_i^d \in \{1, 2, 3, 4, 5\}$ for each dimension $d \in \{C, S, H\}$:
\begin{equation}
    (r_i^d, \tilde{s}_i^d) = \mathcal{A}_R(x_i, m_i, d).
\end{equation}

To ensure the robustness of the supervision, the agent enters a self-verification stage via the verification module $\mathcal{A}_V$ to examine internal consistency across dimensions. 
For example, high cellularity with low clarity may indicate a poorly focused region, prompting the agent to revisit its initial assessment. The verification module then jointly considers the image, preliminary rationales, and actions to resolve inconsistencies and produce the final quality score vector:
\begin{equation}
    \mathbf{s}_i = \mathcal{A}_V \left( x_i, \{(r_i^d, \tilde{s}_i^d)\}_{d \in \{C, S, H\}} \right),
\end{equation}
where $\mathbf{s}_i = [s_i^C, s_i^S, s_i^H]^\top$. This closed-loop mechanism reinforces reliability, yielding an expert-supervised pilot set $\mathcal{D}_P = \{(x_i, \mathbf{s}_i)\}_{i=1}^N$, where $N$ is the number of sampled images.

\subsubsection{Learning the Quality Regressor}
Applying our agent directly to the massive pool $\mathcal{D}_U$ incurs prohibitive computational overhead. To overcome this, we adopt a semi-supervised paradigm, training a lightweight convolutional neural network (\textit{i.e.}, ResNet-50~\cite{he2016deep}) as a quality regressor, denoted as $f_\theta$, to approximate the expert agent.

Using the labeled pilot set $\mathcal{D}_P$, the regressor maps a raw input image $x_i$ to continuous predicted scores $\hat{\mathbf{s}}_i = f_{\theta}(x_i)$. The model is optimized to minimize the discrepancy between its predictions and the agent's scores using the Mean Squared Error (MSE) objective:

\begin{equation}
\mathcal{L}_{\text{reg}} = \frac{1}{N}\sum_{i=1}^{N}\left\| f_{\theta}(x_i)-\mathbf{s}_i \right\|_2^2.
\end{equation}
Once optimized, $f_{\theta}$ can be deployed for real-time quality estimation across the entire target pool $\mathcal{D}_U$, preserving expert-level accuracy while bypassing the Pathology Expert Agent's inference latency.

\subsubsection{Scoring and Quantile-based Selection}
\begin{figure}[t]
    \centering               
\includegraphics[width=0.46 \textwidth]{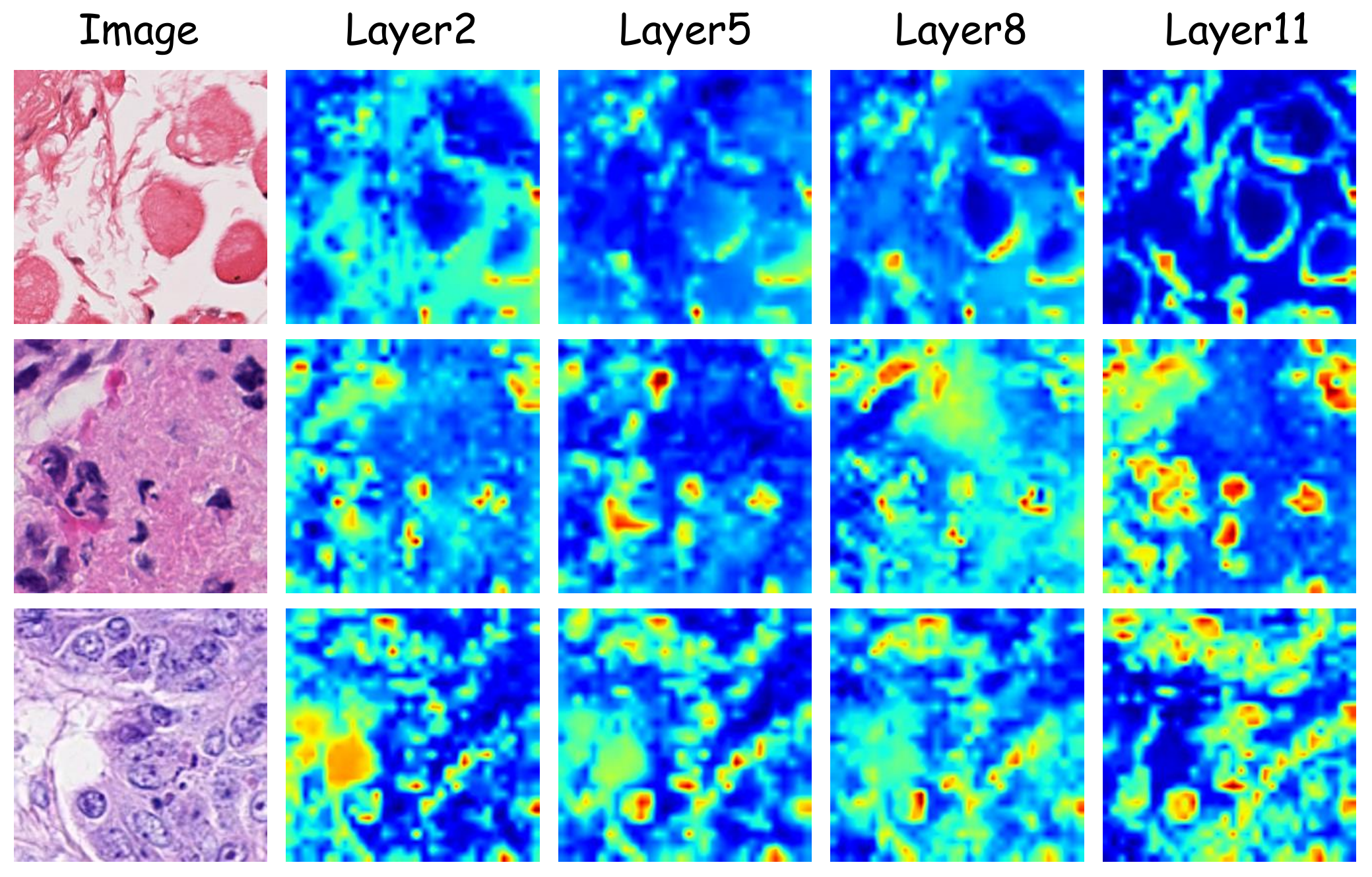}
    \caption{Visualizing the refinement of feature activations in the DiT generative prior. Moving from shallow to deep layers, the model sequentially learns to disentangle semantic structures at various scales, such as individual cells, glandular tissues, and the surrounding background, which provides a crucial foundation for robust semantic segmentation.}
    \label{fig:activation}
\end{figure}
We deploy the trained regressor $f_\theta$ to evaluate all images in the unannotated pool $\mathcal{D}_U$. For each $x_j \in \mathcal{D}_U$, we obtain the predicted multi-dimensional scores:
\begin{equation}
\hat{\mathbf{s}}_j = [ \hat{s}_j^C, \hat{s}_j^S, \hat{s}_j^H ]^\top.
\end{equation}
To establish a unified ranking criterion, we compute the empirical score distributions for each quality dimension across the entire target pool $\mathcal{D}_U$. Rather than relying on rigid absolute thresholds, which can be sensitive to domain shifts, we implement a quantile-based selection strategy. Specifically, let $\tau_d$ denote the score threshold corresponding to the top $k\%$ of the distribution for dimension $d \in \{C, S, H\}$. A patch is selected if and only if its predicted scores all satisfy this top $k\%$ criterion across all three critical dimensions. Formally, the filtered subset $\mathcal{D}^*$ is defined as:

\begin{equation}
\mathcal{D}^* = \left\{ x_j \in \mathcal{D}_U \mid \hat{s}_j^C \ge \tau_C \land \hat{s}_j^S \ge \tau_S \land \hat{s}_j^H \ge \tau_H \right\}.
\end{equation}

This strict filtering ensures that no single severe defect (e.g., severe out-of-focus blur, despite adequate cellularity) compromises the overall utility of the image. Following this process, we distill the massive unannotated pool down to a highly refined subset comprising 217,030 premium-quality patches, which we designate as \textbf{TCGA-200K}. Quantitative analysis reveals a substantial enhancement in data quality: compared to the unfiltered pool $\mathcal{D}_U$, as shown in Fig.~\ref{fig:path_score}, the average predicted scores in TCGA-200K improved by approximately 48.7\% (3.065 $\rightarrow$ 4.558), 22.4\% (3.631 $\rightarrow$ 4.446), and 21.4\% (3.607 $\rightarrow$ 4.379) across the cell presence, histologic clarity, and stain quality dimensions, respectively.

\begin{figure}[t]
    \centering               
\includegraphics[width=0.46 \textwidth]{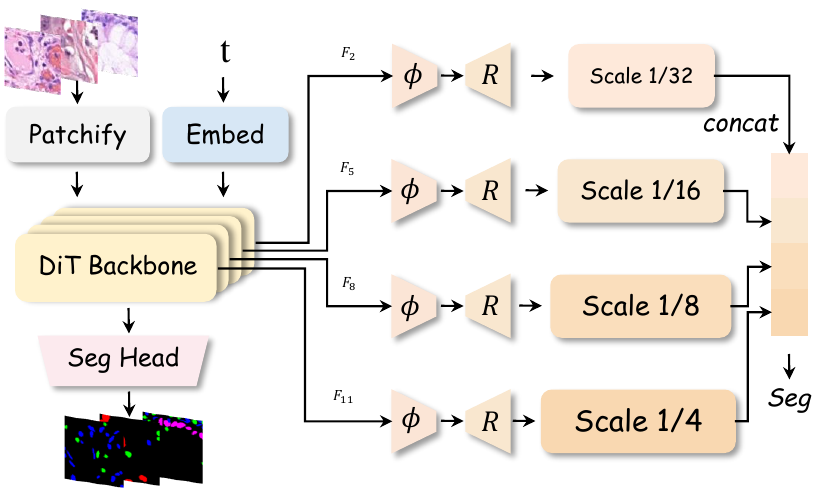}
\caption{Stage III: Transferring generative priors to downstream segmentation. Multi-level features are extracted from the pathology-pretrained DiT backbone, transformed and fused into a unified representation, and then fed into the segmentation head.}
    \label{fig:seghead}
\end{figure}

\subsection{Stage II: Pathology Generative Priors Pretraining}
\newcommand{\pmstd}[2]{%
  \ensuremath{#1_{\scriptscriptstyle \pm #2}}%
}

\newcommand{\bestpmstd}[2]{%
  \ensuremath{\mathbf{#1}_{\scriptscriptstyle \mathbf{\pm #2}}}%
}

\begin{table*}[t]
\centering
\caption{Quantitative comparison on PanNuke and PathSeg-380K. \textsuperscript{*} denotes a statistically significant improvement over the 
corresponding second-best result ($p<0.05$).}
\label{tab:quantitative_comparison}

\small
\setlength{\tabcolsep}{5.0pt}
\renewcommand{\arraystretch}{1.08}

\begin{tabular}{l|cccc|cccc}
\toprule

& \multicolumn{4}{c|}{\textbf{PanNuke}}
& \multicolumn{4}{c}{\textbf{PathSeg-380K}} \\

\cmidrule(lr){2-5}
\cmidrule(lr){6-9}

\textbf{Method}
& \textbf{mIOU (\%)}$\uparrow$
& \textbf{mDICE (\%)}$\uparrow$
& \textbf{p-Acc (\%)}$\uparrow$
& \textbf{mASD}$\downarrow$
& \textbf{mIOU (\%)}$\uparrow$
& \textbf{mDICE (\%)}$\uparrow$
& \textbf{p-Acc (\%)}$\uparrow$
& \textbf{mASD}$\downarrow$ \\

\midrule
\multicolumn{9}{c}{\textit{Task-specific Segmentation Methods}} \\
\midrule

U-Net~\cite{ronneberger2015u}
& \pmstd{48.09}{0.25}
& \pmstd{61.44}{0.42}
& \pmstd{91.56}{0.14}
& \pmstd{29.71}{0.56}
& \pmstd{48.62}{0.94}
& \pmstd{60.84}{0.70}
& \pmstd{92.65}{0.19}
& \pmstd{6.40}{0.15} \\

TransUNet~\cite{chen2021transunet}
& \pmstd{46.63}{1.01}
& \pmstd{58.23}{0.93}
& \pmstd{91.88}{0.28}
& \pmstd{23.05}{0.67}
& \pmstd{46.68}{0.96}
& \pmstd{59.27}{0.71}
& \pmstd{92.29}{0.21}
& \pmstd{10.33}{0.19} \\

HoVer-Net~\cite{graham2019hover}
& \pmstd{51.38}{0.35}
& \pmstd{64.83}{0.54}
& \pmstd{91.74}{0.20}
& \pmstd{22.42}{0.93}
& \pmstd{52.41}{0.75}
& \pmstd{65.73}{0.56}
& \pmstd{92.56}{0.16}
& \pmstd{7.89}{0.13} \\

CPP-Net~\cite{chen2023cpp}
& \pmstd{47.58}{0.17}
& \pmstd{60.82}{0.21}
& \pmstd{91.53}{0.14}
& \pmstd{29.45}{0.28}
& \pmstd{50.17}{0.77}
& \pmstd{62.21}{0.57}
& \pmstd{92.73}{0.16}
& \pmstd{7.58}{0.15} \\

CellViT~\cite{horst2024cellvit}
& \pmstd{51.54}{0.47}
& \pmstd{65.13}{0.59}
& \pmstd{92.14}{0.05}
& \pmstd{26.60}{0.79}
& \pmstd{47.22}{0.83}
& \pmstd{61.26}{0.61}
& \pmstd{91.23}{0.22}
& \pmstd{8.51}{0.33} \\

Un-SAM~\cite{chen2024sam}
& \pmstd{52.31}{0.68}
& \pmstd{65.88}{0.93}
& \pmstd{92.23}{0.11}
& \pmstd{26.47}{0.92}
& \pmstd{52.85}{0.92}
& \pmstd{66.33}{0.68}
& \pmstd{91.81}{0.19}
& \pmstd{5.78}{0.20} \\

\midrule
\multicolumn{9}{c}{\textit{Pretrained Representation-based Methods}} \\
\midrule

SegDINO~\cite{yang2025segdino}
& \pmstd{51.62}{0.52}
& \pmstd{65.17}{0.75}
& \pmstd{91.65}{0.06}
& \bestpmstd{21.75}{0.61}
& \pmstd{49.43}{0.90}
& \pmstd{61.97}{0.66}
& \pmstd{92.74}{0.20}
& \pmstd{12.80}{0.18} \\

MAE~\cite{he2022masked}
& \pmstd{50.54}{0.33}
& \pmstd{63.84}{0.43}
& \pmstd{91.42}{0.15}
& \pmstd{25.34}{0.50}
& \pmstd{50.20}{0.13}
& \pmstd{63.34}{0.96}
& \pmstd{91.36}{0.15}
& \pmstd{7.95}{0.26} \\

CONCH~\cite{lu2024visual}
& \pmstd{53.39}{0.26}
& \pmstd{66.73}{0.44}
& \pmstd{91.89}{0.20}
& \pmstd{22.45}{0.65}
& \pmstd{53.74}{0.75}
& \pmstd{66.32}{0.72}
& \pmstd{91.77}{0.21}
& \pmstd{6.72}{0.16} \\

UNI~\cite{chen2024towards} 
& \pmstd{53.77}{0.59}
& \pmstd{66.70}{0.82}
& \pmstd{92.25}{0.18}
& \pmstd{22.26}{0.48}
& \pmstd{54.59}{0.77}
& \pmstd{65.18}{0.36}
& \pmstd{92.49}{0.12}
& \pmstd{5.37}{0.14} \\

\midrule

Ours
& $\mathbf{55.46}_{\scriptscriptstyle \pm 0.34}^{*}$
& $\mathbf{68.81}_{\scriptscriptstyle \pm 0.39}^{*}$
& $\mathbf{92.77}_{\scriptscriptstyle \pm 0.14}^{*}$
& $22.91_{\scriptscriptstyle \pm 0.56}$
& $\mathbf{55.45}_{\scriptscriptstyle \pm 0.71}$
& $\mathbf{67.81}_{\scriptscriptstyle \pm 0.52}$
& $\mathbf{93.44}_{\scriptscriptstyle \pm 0.16}^{*}$
& $\mathbf{4.81}_{\scriptscriptstyle \pm 0.11}^{*}$ \\

\bottomrule
\end{tabular}
\end{table*}
Using the high-quality TCGA-200K dataset curated in Stage I, Stage II focuses on learning a strong pathology generative prior. 
We adopt a Diffusion Transformer (DiT) architecture~\cite{peebles2023scalable} optimized via a flow matching objective~\cite{lipman2022flow}. 
As shown in Fig.~\ref{fig:dit}, let $x_0$ denote the image from TCGA-200K, and $x_1 \sim \mathcal{N}(0, \mathbf{I})$ denote the Gaussian noise. We define an interpolation path to formulate the intermediate state $x_t$ at timestep $t \in [0, 1]$:
\begin{equation}
x_t = (1 - t) x_0 + t x_1.
\end{equation}
The corresponding target vector field driving this transformation is defined as $v = x_1 - x_0$. The DiT network, parameterized by $\phi$, is optimized to predict this vector field $\hat{v}_\phi(x_t, t)$ by minimizing the Mean Squared Error (MSE) loss:
\begin{equation}
\mathcal{L}_{FM} = \mathbb{E}_{x_0, x_1, t} [\|\hat{v}_\phi(x_t, t) - (x_1 - x_0)\|_2^2]. \label{loss:loss_fm}
\end{equation}
Throughout the network, the timestep $t$ is embedded and injected into the Transformer blocks via Adaptive Layer Normalization (AdaLN~\cite{peebles2023scalable}). 

Once trained, the model generates a pathology image by sampling
$x_1\sim\mathcal{N}(0,\mathbf{I})$ and solving the learned ordinary
differential equation backward from $t=1$ to $t=0$:
\begin{equation}
\frac{\mathrm{d}x_t}{\mathrm{d}t}
=
\hat{v}_{\phi}(x_t,t),
\qquad
\widetilde{x}_0
=
x_1-\int_{0}^{1}\hat{v}_{\phi}(x_t,t)\,\mathrm{d}t.
\label{eq:fm_sampling}
\end{equation}

By training on the patches selected in Stage I, the model learns the distribution of informative pathology regions. As shown in Fig.~\ref{fig:fake_img}, the generated samples closely resemble real pathology images in both cellular morphology and tissue organization, a result further supported by the quantitative and expert evaluations presented in later experiments.

\subsection{Stage III: Generative Features for Segmentation}
In Stage III, we transfer the generative representations learned by the DiT to pathology semantic segmentation. Related studies have shown that intermediate features from pretrained diffusion models can support dense prediction~\cite{baranchuk2021label,wu2024medsegdiff}. Different from these methods, our backbone is pretrained on the agent-curated pathology dataset and therefore captures pathology-specific cellular morphology and organization.

As shown in Fig.~\ref{fig:seghead}, given an input image $x$, we first construct an intermediate state using the same interpolation path as in Stage II:
\begin{equation}
x_t = (1-t)x + t\epsilon,
\qquad
\epsilon \sim \mathcal{N}(0,\mathbf{I}),
\end{equation}
and feed $(x_t,t)$ into the pretrained DiT $v_\phi$. The timestep $t$ determines the corruption level of the input and consequently affects the balance between local appearance and high-level semantic information in the extracted features. We treat $t$ as a hyperparameter and study its effect in Sec.~\ref{sec:ablation_t}.

Inspired by the multi-level feature extraction strategy of DPT~\cite{ranftl2021vision}, we extract representations from DiT blocks 2, 5, 8, and 11, which span different depths of the backbone. Since the DiT operates on flattened token sequences, these representations are rearranged into 2D feature maps, denoted as $\{F_k\}_{k=1}^{4}$. A lightweight multi-level decoder then projects them into hierarchical spatial resolutions.
Specifically, each $F_k$ is processed by a scale-specific projection $\phi_k(\cdot)$ consisting of channel reduction and strided convolutions. The projected features are resized to a common spatial resolution by bilinear interpolation $\mathrm{R}(\cdot)$ and refined by a convolutional block $\mathcal{C}_k(\cdot)$:
\begin{equation}
\widetilde{F}_k
=
\mathcal{C}_k
\left(
\mathrm{R}\left(\phi_k(F_k)\right)
\right),
\qquad k=1,\ldots,4.
\end{equation}
The aligned features are concatenated and passed to the segmentation head $\mathcal{H}_{\mathrm{seg}}$ to obtain the pixel-wise prediction:
\begin{equation}
M =
\mathcal{H}_{\mathrm{seg}}
\left(
\operatorname{Concat}
\left[
\widetilde{F}_1,
\widetilde{F}_2,
\widetilde{F}_3,
\widetilde{F}_4
\right]
\right).
\end{equation}

The segmentation network is optimized using a combination of cross-entropy and Dice losses~\cite{milletari2016v}:
\begin{equation}
\mathcal{L}_{\mathrm{seg}}
=
\mathcal{L}_{\mathrm{CE}}
+
\mathcal{L}_{\mathrm{Dice}}.
\label{loss:loss_ce}
\end{equation}

As shown in Fig.~\ref{fig:activation}, the multi-level DiT features capture both local cellular structures and broader tissue context. The segmentation decoder therefore focuses on integrating these representations and mapping them to task-specific labels, rather than relearning pathology morphology from limited annotations. This allows the learned generative prior to be effectively transferred to semantic segmentation.
\section{Experiments}
\begin{figure*}[t]
    \centering               
\includegraphics[width=0.9\textwidth]{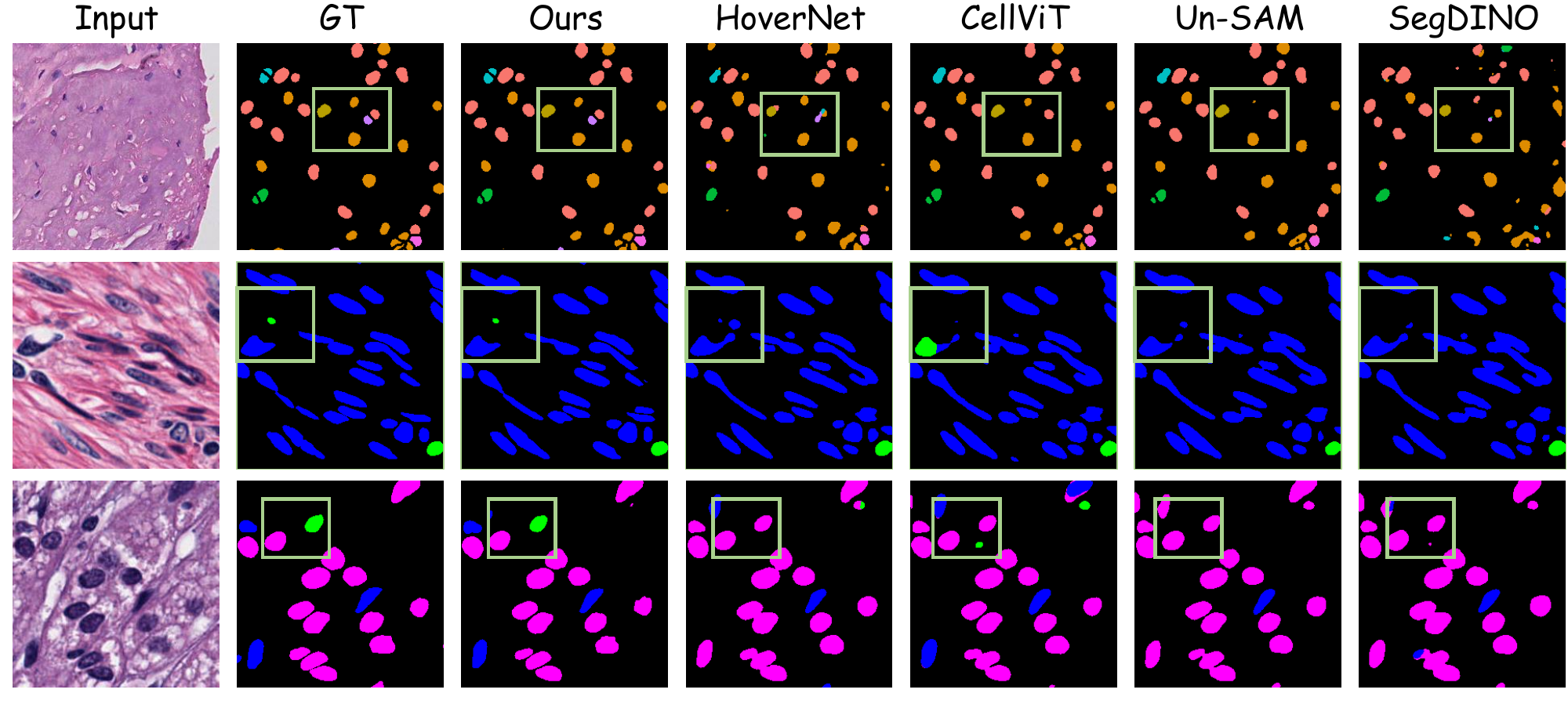}
\caption{Visual comparison of segmentation results across pathology tissue types. Green boxes highlight challenging regions where our framework better matches the ground truth (GT), particularly under cluttered backgrounds and varying cell scales.}
\label{fig:visual comparisons}
\end{figure*}
\subsection{Datasets}
As shown in Fig.~\ref{fig:dataset}(b), we use two pathology segmentation datasets for downstream evaluation: the public PanNuke dataset~\cite{gamper2019pannuke} and our proposed PathSeg-380K dataset. PanNuke contains 5 nucleus categories: Inflammatory (32,276), Connective (50,585), Neoplastic (77,403), Non-neoplastic epithelial (26,572), and Dead (2,908). PathSeg-380K is our in-house pathology segmentation dataset, collected with ethical approval, covering 12 categories: Mast (3,253), DC (9,087), pDC (15,550), B (44,993), T (42,944), Endothelial (49,498), Plasma (82,850), Mono/Mac (59,319), Fibroblast (69,868), MSC (114), Cycling (1,436), and NA (2,720). Together, they cover diverse semantic categories and morphological patterns for evaluating segmentation performance.

\subsection{Implementation Details}
In Stage II, we train the pathology generative prior using a DiT backbone with the objective defined in Equation~\ref{loss:loss_fm}. The model takes $256\times256$ patches as input, featuring a patch size of $8$, an embedding dimension of 768, 12 Transformer blocks, and 12 attention heads. We employ the AdamW optimizer with a learning rate of $1\times10^{-4}$ and a weight decay of $1\times10^{-4}$. The model is trained for 200,000 iterations with a total batch size of 16. In Stage III, the segmentation framework is initialized with the DiT backbone pretrained in Stage II. We optimize the network using AdamW, explicitly applying a learning rate of $1 \times 10^{-5}$ to the backbone and $3 \times 10^{-4}$ to the newly initialized segmentation head. To ensure fair evaluation, we employ a 3-fold cross-validation strategy, training the model for 20 epochs per fold using the loss function defined in Eq.~\ref{loss:loss_ce}. All experiments are conducted in parallel across four NVIDIA A6000 GPUs.

\subsection{Reliability of Agent-based Quality Assessment}
We implement the pathology expert agent using GPT-4.1 mini (``gpt-4.1-mini-2025-04-14'') and evaluate its scoring reliability on an independent set of 600 pathology patches, which are excluded from the pilot set used to train the quality regressor. Two pathologists independently assess each patch in terms of cell presence, stain quality, and histologic clarity. Samples with a score difference of two or more points are jointly reviewed to obtain the final consensus score. We use the pathologist consensus as the reference and report Spearman's correlation~\cite{spearman1961proof} for each quality dimension. Mean absolute error (MAE) and quadratic weighted Cohen's $\kappa$~\cite{cohen1968weighted} are averaged across the three dimensions.

As shown in Table~\ref{tab:agent_reliability}, the two pathologists show consistent assessments, with Spearman's correlations above 0.81 and an average weighted $\kappa$ of 0.827. The agent reaches correlations of 0.782, 0.811, and 0.829 for cell presence, stain quality, and histologic clarity, respectively. Its average MAE is 0.326, while the weighted $\kappa$ against the pathologist consensus is 0.814, close to the agreement between the two pathologists. The quality regressor shows a small decrease against the consensus, with an MAE of 0.378 and a weighted $\kappa$ of 0.789. Meanwhile, its correlations with the agent are all above 0.90, with an MAE of 0.170 and a weighted $\kappa$ of 0.912. These results indicate that the regressor closely reproduces the agent scores while retaining strong agreement with pathologist assessment, supporting its use for large-scale patch screening.

\begin{table}[t]
\centering
\caption{Reliability of pathology quality assessment against
pathologist consensus on an independent test set.}
\label{tab:agent_reliability}
\scriptsize
\setlength{\tabcolsep}{3.8pt}
\renewcommand{\arraystretch}{1.08}

\begin{tabular}{lccc|cc}
\toprule
\textbf{Evaluator}
& \multicolumn{3}{c|}{\textbf{Spearman's $\rho$} $\uparrow$}
& \textbf{MAE} $\downarrow$
& $\boldsymbol{\kappa}$ $\uparrow$ \\
\cmidrule(lr){2-4}
& \textbf{Cell}
& \textbf{Stain}
& \textbf{Clarity}
& \textbf{Avg.}
& \textbf{Avg.} \\
\midrule

Pathologist 1 vs. Pathologist 2
& 0.815 & 0.835 & 0.816
& 0.306 & 0.827 \\

Agent vs. Consensus
& 0.782 & 0.811 & 0.829
& 0.326 & 0.814 \\

Regressor vs. Consensus
& 0.759 & 0.797 & 0.796
& 0.378 & 0.789 \\

Regressor vs. Agent
& 0.909 & 0.914 & 0.913
& 0.170 & 0.912 \\

\bottomrule
\end{tabular}
\end{table}
\subsection{Effect of Agent-based Data Curation}
We evaluate generation quality using FID~\cite{heusel2017gans}, KID~\cite{sutherland2018demystifying}, and Mean Opinion Score (MOS), where lower FID and KID and higher MOS indicate better quality. MOS is independently assessed by two pathologists on a five-point scale, with ``1'' and ``5'' indicating the lowest and highest quality, respectively, and their average score is reported. For each curation setting, we train the generative model on the corresponding retained subset, generate 10,000 pathology images, and compute FID and KID against real images. Downstream utility is evaluated by transferring the learned backbone to PanNuke under the same segmentation protocol, with mIoU as the metric. In Table~\ref{tab:agent_curation}, \textit{From Scratch} denotes the full segmentation architecture with a randomly initialized DiT backbone and the same multi-level decoder, while \textit{w/o Agent (100\%)} denotes generative pretraining on the complete unfiltered training set. For the remaining settings, an image is retained only when its predicted scores for cell presence, stain quality, and histologic clarity simultaneously fall within the top 75\%, 50\%, 25\%, 10\%, or 5\% of their respective distributions.
\begin{table}[t]
\centering
\caption{
Effect of agent-based data curation on generation quality and
downstream segmentation performance on PanNuke.
}
\label{tab:agent_curation}

\scriptsize
\setlength{\tabcolsep}{3.2pt}
\renewcommand{\arraystretch}{1.08}

\begin{tabular}{@{}lcccc@{}}
\toprule
\textbf{Curation Setting}
& \textbf{FID} $\downarrow$
& \textbf{KID} $\downarrow$
& \textbf{MOS} $\uparrow$
& \textbf{mIoU (\%)} $\uparrow$ \\
\midrule

From Scratch
& -- & -- & -- & 47.19 \\

w/o Agent (100\%)
& 38.75
& 0.056
& 3.12
& 53.41 \\

w/ Agent (Top 75\%)
& 33.61
& 0.032
& 4.06
& 54.42 \\

w/ Agent (Top 50\%)
& 30.08
& 0.028
& 3.87
& 54.46 \\

w/ Agent (Top 25\%)
& 24.94
& 0.024
& 4.20
& 55.12 \\

\textbf{w/ Agent (Top 10\%)}
& 23.96
& 0.021
& \textbf{4.43}
& \textbf{55.46} \\

w/ Agent (Top 5\%)
& \textbf{23.11}
& \textbf{0.019}
& 4.41
& 55.03 \\

\bottomrule
\end{tabular}
\end{table}
Using the complete unfiltered dataset yields an FID of 38.75, a KID of 0.056, and an MOS of 3.12. Agent-based curation generally improves generation quality as the selection criterion becomes stricter. In particular, requiring samples to rank within the top 10\% in each quality dimension reduces FID and KID to 23.96 and 0.021, respectively, while achieving the highest MOS of 4.43. Further tightening the threshold to the top 5\% improves FID and KID to 23.11 and 0.019, but slightly decreases MOS to 4.41. These results indicate that training with higher-ranked samples improves distributional fidelity and pathological plausibility.

The downstream results show a similar trend. Training the segmentation model from scratch achieves an mIoU of 47.19\%, while transferring the generative prior trained without agent-based curation increases it to 53.41\%. Using the prior trained on samples ranked within the top 10\% in each quality dimension further improves mIoU to 55.46\%, corresponding to gains of 8.27 and 2.05 percentage points over the two settings, respectively. Although the top 5\% setting achieves the best FID and KID, its mIoU decreases to 55.03\%, suggesting that overly strict filtering may reduce the diversity needed for transferable representation learning. We therefore adopt the top 10\% setting in subsequent experiments.

\subsection{Comparison on Downstream Segmentation}
\subsubsection{Quantitative Comparisons}
We compare our framework with two groups of methods. The first group consists of task-specific segmentation models, including CNN-based U-Net~\cite{ronneberger2015u}, HoVer-Net~\cite{graham2019hover}, and CPP-Net~\cite{chen2023cpp}; Transformer-based TransUNet~\cite{chen2021transunet} and CellViT~\cite{horst2024cellvit}; and SAM-based Un-SAM~\cite{chen2024sam}. The second group evaluates pretrained encoders coupled with the same segmentation decoder as our framework, including SegDINO~\cite{caron2021emerging,yang2025segdino} and MAE~\cite{he2022masked} pretrained on natural images, as well as CONCH~\cite{lu2024visual} and UNI~\cite{chen2024towards} pretrained on pathology images.

As shown in Table~\ref{tab:quantitative_comparison}, our framework achieves the best overlap-based segmentation performance on PanNuke, with an mIoU of 55.46\%, an mDice of 68.81\%, and a p-Acc of 92.77\%. Compared with the corresponding second-best results, it improves mIoU, mDice, and p-Acc by 1.69, 2.08, and 0.52 percentage points, respectively. On PathSeg-380K, our framework achieves the best results across all four metrics, reaching 55.45\% mIoU, 67.81\% mDice, 93.44\% p-Acc, and 4.81 mASD. It surpasses the corresponding second-best results by 0.86, 1.48, and 0.70 percentage points in mIoU, mDice, and p-Acc, respectively, while reducing mASD from 5.37 to 4.81. Improvements marked with an asterisk in Table~\ref{tab:quantitative_comparison} are statistically significant compared with the corresponding second-best method under a paired two-sided $t$-test ($p<0.05$). These results demonstrate that the transferred generative representations improve segmentation accuracy across both datasets, particularly in terms of region overlap.

\subsubsection{Visual Comparisons}
As shown in Fig.~\ref{fig:visual comparisons}, our method produces segmentation results that better match the ground truth. In challenging regions containing small or densely clustered nuclei, it reduces missed predictions and boundary errors while providing more complete masks. These observations are consistent with the quantitative results in Table~\ref{tab:quantitative_comparison}.

\begin{table}[t]
\centering
\caption{Ablation study of the proposed components.}
\label{tab:module_ablation}

\resizebox{0.48\textwidth}{!}{%
\begin{tabular}{@{\hspace{4pt}}lccc|cc|cc}
\toprule
& \multicolumn{3}{c|}{}
& \multicolumn{2}{c|}{\textbf{PanNuke}}
& \multicolumn{2}{c}{\textbf{PathSeg-380K}} \\
\multirow{-2}{*}{\textbf{Method}}
& \multirow{-2}{*}{\textbf{Gen. Prior}}
& \multirow{-2}{*}{\textbf{Agent}}
& \multirow{-2}{*}{\textbf{Seg. Head}}
& \textbf{mIoU (\%)}$\uparrow$
& \textbf{mASD}$\downarrow$
& \textbf{mIoU (\%)}$\uparrow$
& \textbf{mASD}$\downarrow$ \\
\midrule

Baseline
&  &  &
& 44.58
& 30.93
& 44.23
& 8.86 \\

M1
& \checkmark &  &
& 52.17
& 25.86
& 52.08
& 6.58 \\

M2
& \checkmark & \checkmark &
& 53.85
& 24.42
& 53.72
& 5.37 \\

M3
& \checkmark &  & \checkmark
& 53.41
& 24.01
& 53.19
& 5.51 \\

\midrule
\textbf{Ours (Full)}
& \checkmark & \checkmark & \checkmark
& \textbf{55.46}
& \textbf{22.91}
& \textbf{55.45}
& \textbf{4.81} \\

\bottomrule
\end{tabular}
}
\end{table}
\subsection{Ablation Study}
\subsubsection{Module Ablation}
To assess the contribution of each component, we conduct the ablation study shown in Table~\ref{tab:module_ablation}. The Baseline uses a randomly initialized DiT with an MLP head. Introducing the pretrained generative prior in M1 improves mIoU by 7.59 and 7.85 percentage points on PanNuke and PathSeg-380K, respectively. Starting from M1, M2 and M3 separately introduce Agent Screening and the multi-level decoder. Agent Screening improves mIoU by 1.68 and 1.64 points, while the multi-level decoder yields gains of 1.24 and 1.11 points and reduces mASD from 25.86 to 24.01 on PanNuke and from 6.58 to 5.51 on PathSeg-380K. Combining all components achieves the best performance on both datasets, reaching mIoUs of 55.46\% and 55.45\% and mASDs of 22.91 and 4.81, respectively. These results demonstrate that agent-based data curation and multi-level feature fusion provide complementary benefits to the generative prior.

\subsubsection{Hyperparameter Ablation}
\label{sec:ablation_t}
The timestep $t$ controls the semantic level of features extracted from the DiT backbone. Although the model is trained over the full flow trajectory $t\in[0,1]$, downstream segmentation uses a fixed $t$ without performing iterative denoising. Through AdaLN~\cite{peebles2023scalable}, smaller values emphasize local textures and cellular details, whereas larger values favor more abstract tissue semantics. We therefore investigate the optimal balance between spatial detail and semantic abstraction.
\begin{figure}[t]
    \centering               
\includegraphics[width=0.46 \textwidth]{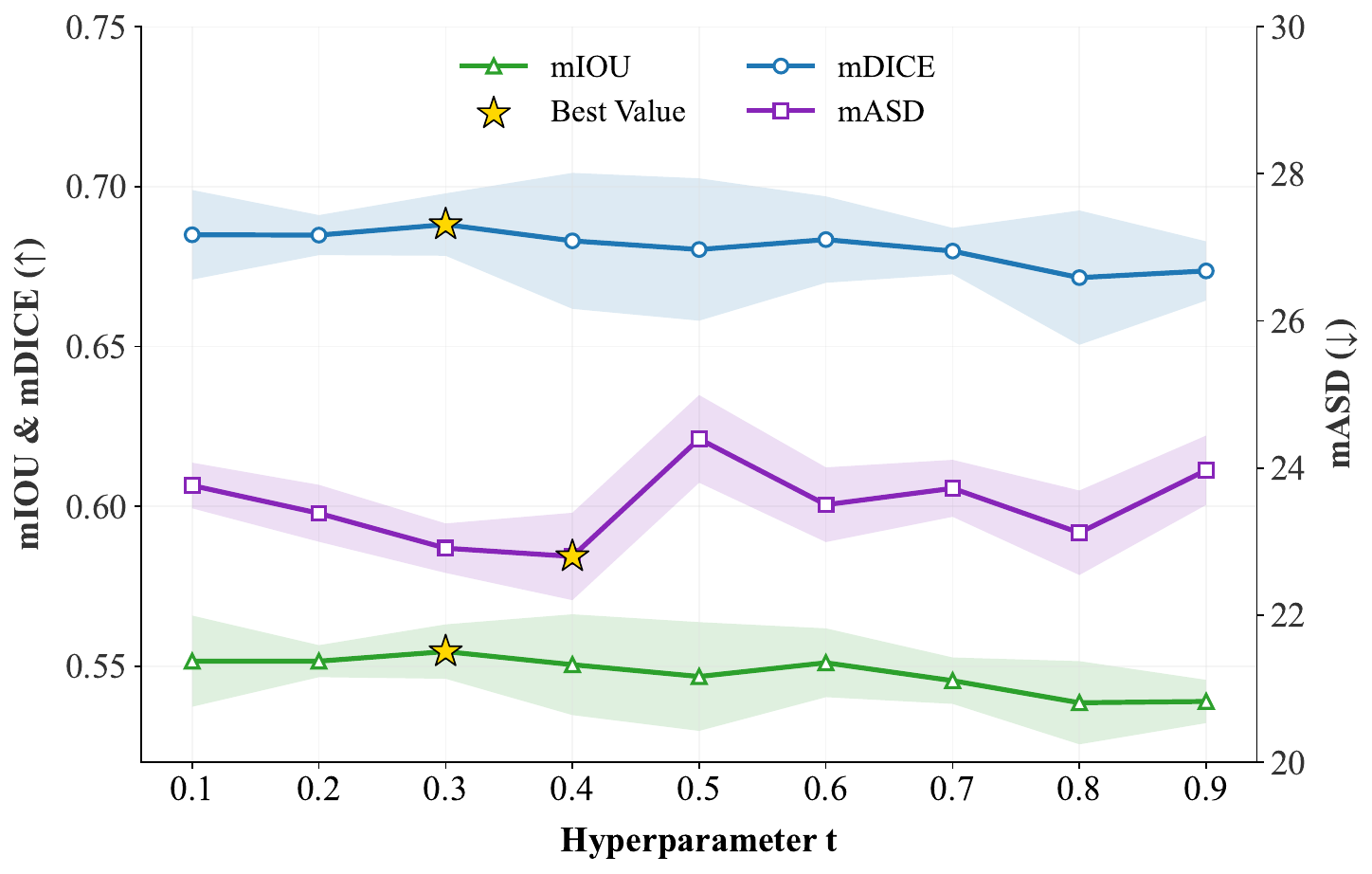}
\caption{Effect of hyperparameter $t$ on mIoU, mDice, and mASD. Yellow stars highlight the best-performing settings for each metric.}
    \label{fig:ablation_t}
\end{figure}
As shown in Fig.~\ref{fig:ablation_t}, increasing $t$ from 0.1 to 0.3 improves the overlap-based metrics, with $t{=}0.3$ achieving the best mIoU of 55.46\% and mDice of 68.81\%, together with a competitive mASD of 22.91. Although $t{=}0.4$ yields a slightly lower mASD, its mIoU and mDice are inferior to those obtained at $t{=}0.3$. At larger timesteps, the overall segmentation performance generally declines, with mIoU decreasing to 53.58\% at $t{=}0.9$, likely because excessive corruption weakens fine-grained spatial information. We therefore set $t{=}0.3$ for all experiments as the best overall trade-off between region overlap and boundary accuracy.

\section{Conclusion}
In this work, we presented an agent-driven framework that integrates pathology data curation, generative prior learning, and downstream semantic segmentation. The pathology expert agent enables scalable quality screening and the construction of the curated TCGA-200K dataset. Using these high-quality samples, the DiT backbone learns pathology-specific representations that capture both cellular morphology and tissue-level semantics, which are effectively transferred to a lightweight segmentation head. Strong results on PanNuke and PathSeg-380K demonstrate the value of generative pretraining for both pathology image synthesis and downstream analysis, while future work will extend the framework to other clinically relevant tasks, including classification and survival analysis.
We will also investigate more efficient agent-based curation strategies to reduce screening costs while preserving data quality across institutions and staining protocols.

\clearpage
\newpage
\bibliographystyle{unsrt}
\bibliography{ref}

\end{document}